\documentclass[sigconf, nonacm]{acmart}

\usepackage{booktabs}
\usepackage{multirow}
\usepackage{array}
\usepackage{xcolor}
\usepackage{colortbl}
\usepackage{xspace}
\usepackage{adjustbox}

\newcommand\vldbyear{2026}
\newcommand\vldbworkshop{Tabular Data Analysis (TaDA)}
\newcommand\vldbauthors{\authors}
\newcommand\vldbtitle{\shorttitle}
\newcommand\vldbavailabilityurl{}
\newcommand\vldbpagestyle{plain}

\newcommand{\plurel}{\textsc{PluRel}\xspace}
\newcommand{\rdbpfn}{\textsc{RDB-PFN}\xspace}
\newcommand{\relsalt}{\textsc{Rel-SALT}\xspace}
\newcommand{\dfs}{\textsc{DFS}\xspace}

\begin{document}
\title[Curriculum Matters: Data-Efficient Relational PFN Pretraining]{Curriculum Matters: Data-Efficient Relational PFN Pretraining with Synthetic Data}

% %%
% %% Authors, update with actual author information
% \author{Anonymous Author(s)}
% \affiliation{%
%   \institution{Anonymous Institution}
%   \city{Anonymous City}
%   \country{Anonymous Country}
% }
% \email{anonymous@example.com}

\author{Mohammad Sadeq Abolhasani}
\affiliation{%
  \institution{SAP Labs, LLC.}
  \city{Palo Alto}
  \state{CA}
  \postcode{94304}
  \country{United States}
}
\email{Rastin.Abolhasani@sap.com}

\author{Viswanath Ganapathy}
\affiliation{%
  \institution{SAP Labs, LLC.}
  \city{Palo Alto}
  \state{CA}
  \postcode{94304}
  \country{United States}
}
\email{Viswa.Ganapathy@sap.com}

%%
%% The abstract is a short summary of the work
\begin{abstract}
Relational Prior-Data Fitted Networks (PFNs) such as \rdbpfn{} approximate Bayesian inference over multi-table relational databases by pretraining on millions of synthetic tasks. We investigate three intertwined questions about this paradigm. First, can a structurally different synthetic generator \plurel{} substitute for \rdbpfn{}'s prior? Second, how much does the \emph{order} in which synthetic data is presented to the PFN affect downstream performance? Third, how much relational reasoning can a PFN acquire from single-table synthetic pretraining alone, before any relational data is introduced? Using \plurel{} as the sole synthetic data source across all experiments, we find: (i) a progressive single-table curriculum that gradually widens schema complexity from 7 to 17 columns reaches 0.703 average ROC-AUC on the 23-task tabular benchmark using only $\sim$13{,}300 synthetic tables (approximately 45$\times$ fewer single-table datasets than \rdbpfn{}'s reported warm-up recipe), while the same data trained all-at-once collapses to 0.541 ROC-AUC; (ii) a relational curriculum trained from scratch on only $\sim$5{,}500 \plurel{} databases reaches 0.638 average ROC-AUC on the 19-task RelBench/4DBInfer benchmark, recovering 88\% of \rdbpfn{}'s reported performance with $\sim$220$\times$ less relational synthetic data; and (iii) the single-table curriculum model, evaluated directly on the relational benchmark without any relational adaptation, achieves 0.631, nearly matching the dedicated relational pipeline. Together, these findings suggest that curriculum design and synthetic data diversity may matter more for relational PFN pretraining than the specific relational generator or raw synthetic scale alone.
\end{abstract}

\maketitle

%%% do not modify the following VLDB block %%
%%% VLDB block start %%%
\pagestyle{\vldbpagestyle}
\begingroup\small\noindent\raggedright\textbf{VLDB Workshop Reference Format:}\\
\vldbauthors. \vldbtitle. VLDB \vldbyear\ Workshop: \vldbworkshop.\\
\endgroup
\begingroup
\renewcommand\thefootnote{}\footnote{\noindent
This work is licensed under the Creative Commons BY-NC-ND 4.0 International License. Visit \url{https://creativecommons.org/licenses/by-nc-nd/4.0/} to view a copy of this license. For any use beyond those covered by this license, obtain permission by emailing \href{mailto:info@vldb.org}{info@vldb.org}. Copyright is held by the owner/author(s). Publication rights licensed to the VLDB Endowment. \\
\raggedright Proceedings of the VLDB Endowment.
ISSN 2150-8097. \\
}\addtocounter{footnote}{-1}\endgroup
%%% VLDB block end %%%

%%% do not modify the following VLDB block %%
%%% VLDB block start %%%
\ifdefempty{\vldbavailabilityurl}{}{
\vspace{.3cm}
\begingroup\small\noindent\raggedright\textbf{VLDB Workshop Artifact Availability:}\\
The source code, data, and/or other artifacts have been made available at \url{\vldbavailabilityurl}.
\endgroup
}
%%% VLDB block end %%%

\section{Introduction}

Relational foundation models lag substantially behind text and vision foundation models in maturity, primarily because high-quality multi-table relational databases (RDBs) are scarce, proprietary, and structurally heterogeneous. Two recent threads address this. \plurel{}\cite{kothapalli2026plurel} provides a lightweight framework for sampling diverse synthetic RDBs from scratch via hierarchical stochastic block models and per-table structural causal models. \rdbpfn{}~\cite{wang2026relational} demonstrates that Prior-Data Fitted Network (PFN) style in-context learning~\cite{muller2022pfn,hollmann2022tabpfn}, originally developed for single-table tasks, can be extended to relational data through Deep Feature Synthesis (\dfs)~\cite{kanter2015dfs} linearization combined with a custom neural relational prior (LayerDAG schema generator, attention-based selective structural causal model, and bidirectional graph neural network for content completion). \rdbpfn{}'s reported recipe consumes roughly 1.8M synthetic tasks divided between single-table warm-up and relational adaptation.

In this paper we ask three coupled questions about the \rdbpfn{} paradigm. \textbf{(Q1)}~Is the reported in-context learning capability tied to \rdbpfn{}'s specific neural generator, or would a structurally different synthetic prior with different generating mechanisms suffice? \textbf{(Q2)}~For a PFN trained on synthetic data, how much does the \emph{order} in which synthetic tasks are presented affect downstream quality, separate from the raw synthetic volume? \textbf{(Q3)}~How much relational reasoning capacity does the PFN acquire from single-table synthetic pretraining alone, before any relational data is introduced?

We address all three questions using \plurel{} as the single synthetic data source: substituting it for \rdbpfn{}'s native generator throughout (Q1), running direct curriculum-vs. all-at-once ablations on both stages (Q2), and evaluating the single-table curriculum checkpoint directly on the relational benchmark without any relational adaptation (Q3). The experiments span seven families (Section~\ref{sec:method}).

The results reposition the conversation about \rdbpfn{}-style relational pretraining. Curriculum ordering, not synthetic generator identity and not raw synthetic scale, is the dominant lever. Specifically:
\begin{sloppypar}
\begin{itemize}
  \item A progressive width curriculum over \plurel{}-generated single-table data (7 to 17 columns, $\sim$13{,}300 datasets with 150 rows each) reaches 0.703 average ROC-AUC on the 23-task tabular benchmark, recovering 88\% of \rdbpfn{}'s reported performance while using approximately 45$\times$ fewer datasets and roughly 270$\times$ less total tabular content than the original single-table warm-up; the same data trained all-at-once collapses to 0.541, a 16-point absolute gap (Section~\ref{sec:single}).
  \item A from-scratch relational curriculum over $\sim$5{,}500 \plurel{} databases reaches 0.638 average ROC-AUC on the 19-task relational benchmark, recovering 88\% of \rdbpfn{}'s reported 0.725 with $\sim$220$\times$ less relational synthetic data (Section~\ref{sec:relational}).
  \item The single-table curriculum model, evaluated directly on the 19 relational tasks without any relational adaptation, achieves 0.631, within 1 point of the dedicated relational pipeline (Section~\ref{sec:transfer}).
  \item Ablating either the single-table warm-up or relational curriculum degrades performance, with the warm-up emerging as the more critical stage (Section~\ref{sec:ablations}).
\end{itemize}
\end{sloppypar}

\section{Background}

\subsection{\rdbpfn{} and DFS-Linearized Relational ICL}

\rdbpfn{}~\cite{wang2026relational} is a Prior-Data Fitted Network for relational tasks. It addresses the heterogeneity of multi-table inputs by applying Deep Feature Synthesis as a deterministic linearization that produces a single feature matrix per target entity. Aggregations such as Mean, Max, Min, Count, and Mode summarize neighboring tables at 1-hop and 2-hop depths, after which a compact bidirectional transformer ($\sim$0.7M parameters, 6 layers, $d_{\text{model}}=128$) performs in-context learning with a binary classification head. Training proceeds in a two-stage curriculum: a single-table warm-up on roughly 600{,}000 synthetic tabular tasks, followed by a relational adaptation phase on roughly 1.2 million additional synthetic RDBs.

\subsection{\plurel{} as a Synthetic Generator}

\plurel{}~\cite{kothapalli2026plurel} factorizes relational database generation into three stages. A schema graph is sampled from a family of random directed acyclic graphs (Barab\'asi-Albert, Reverse Random-Tree, Watts-Strogatz). Primary-foreign key connectivity is generated using a hierarchical stochastic block model. Feature columns within each table are generated by structural causal models whose source nodes incorporate temporal patterns. The pipeline is CPU-only and embarrassingly parallel.

\plurel{}'s generating mechanisms are structurally orthogonal to \rdbpfn{}'s native generator. \rdbpfn{} uses a learned LayerDAG schema model, attention-based selective parent assignment, and a global bidirectional GNN over the instance graph for content completion. \plurel{} uses random-graph priors, block-model connectivity, and table-local SCMs without any cross-table message passing during generation. If both lead to comparable PFN performance, the result is informative about which properties of the synthetic prior actually matter.

\subsection{\relsalt{}: Semantic Relational Schema}

\relsalt{}~\cite{ranjan2025rt} extends RelBench ~\cite{robinson2024relbench} with a semantically grounded schema family used by the Relational Transformer. We use it as a seed schema set in the relational stage of our curriculum, mixed with \plurel{}-generated random schema, to ground the model in topologies aligned with real enterprise RDBs.

\section{Methodology}
\label{sec:method}

\plurel{} is the sole synthetic data generator. We organize experiments into seven families that collectively address Q1-Q3.

\paragraph{Family A -- Single-table progressive curriculum.} Starting from random initialization, we train on \plurel-generated single tables in width-progressive stages. The first stage exposes the model to 3{,}300 tables with 7 columns (denoted TF07). Each subsequent stage continues from the previous checkpoint and introduces progressively wider schemas, ranging from TF08 through TF17, with most stages contributing 1{,}000 additional tables (and TF12 contributing 2{,}000). All tables contain 150 rows. The cumulative warm-up corpus comprises approximately 13{,}300 datasets, compared to roughly 600{,}000 in the original \rdbpfn{} recipe.

\paragraph{Family B: Single-table all-at-once ablation.} An ablation using exactly the same TF07-TF17 \plurel{} corpus but mixed into a single training run with no curriculum ordering. This isolates the contribution of curriculum ordering from the contribution of data content.

\paragraph{Family C: Relational curriculum from scratch.} Starting from random initialization (no single-table warm-up), we apply a four-stage relational curriculum: \relsalt{} (500 DBs), then $+$ \plurel{} Small 1 (3,000 DBs), then $+$ \plurel{} Small2 (1,000 DBs), then $+$ \plurel{} Large 1 (1,000 DBs). Each stage continues from the previous checkpoint.

\paragraph{Family D: Single-table model evaluated on relational tasks.} The final TF07-TF17 single-table checkpoint from Family A is evaluated directly on the 19 relational tasks without any further relational adaptation. This probes how much relational capability emerges from \dfs-linearized single-table pretraining alone.

\paragraph{Family E: Relational all-at-once after single-table warm-up.} Starting from the Family A checkpoint, we train on the entire \relsalt{} + Small 1 + Small 2 + Large 1 corpus mixed at once, with no relational curriculum ordering. Compared against the Family A + Family C combination, this isolates the contribution of relational curriculum ordering.

\paragraph{Family F: Relational all-at-once without single-table warm-up.} The full relational corpus is mixed at once, starting from random initialization. Compared to Family E, this isolates the contribution of the single-table warm-up.

\paragraph{Family G: Relational-only, no single tables at all.} The full relational corpus was mixed at once; there was no single-table data at any stage.

\paragraph{Family G: Relational curriculum without \relsalt{}.} A relational curriculum identical to Family C but without the \relsalt{} seed stage; only \plurel{}-generated databases are used, with no single-table warm-up.

\paragraph{Linearization and targets.} All synthetic RDBs are linearized via \dfs{} with the same aggregation primitives as the original \rdbpfn{}. We standardize the post-\dfs{} feature width to 30 columns through random feature subsampling, and sample target columns per the original protocol (binary classification via median-binarization for numeric, one-vs-rest for categorical, 6 random target columns per schema).

\paragraph{Evaluation.} We use the \rdbpfn{} consumer architecture and training code, modifying only the synthetic data source. All other hyperparameters follow the original paper. Single-table evaluation uses 23 classification tasks from the Grinsztajn et al.~\cite{grinsztajn2022treebased} benchmark; relational evaluation uses 19 tasks from RelBench~\cite{robinson2024relbench} and 4DBInfer~\cite{wang20244dbinfer}. We report ROC-AUC at context sizes 64 and 1024 on the relational benchmark, and at context size 1024 on the single-table benchmark.

\section{Results}

\subsection{Single-Table Curriculum (Family A-B)}
\label{sec:single}

Table~\ref{tab:single} reports the average ROC-AUC across the 23 single-table classification tasks at each stage of the width-progressive curriculum. Performance grows monotonically from 0.567 at TF07-only (3{,}300 DBs) to a peak of 0.715 at the TF07-TF12 checkpoint (8{,}300 DBs), then oscillates in the 0.687--0.705 range for subsequent stages, with the final TF07-TF17 checkpoint reaching 0.703. With approximately 45$\times$ less single-table pretraining data, the curriculum recovers 88\% of \rdbpfn{}'s reported 0.800.

\begin{table}[t]
\caption{Single-table benchmark (23 classification tasks, ctx 1024). Width-progressive curriculum over \plurel-generated tables; cumulative dataset count grows from 3{,}300 to 13{,}300.}
\label{tab:single}
\small
\begin{tabular}{lrr}
\toprule
Curriculum stage & Total DBs & Avg.\ ROC-AUC \\
\midrule
TF07               & 3{,}300  & 0.567 \\
$+$ TF08           & 4{,}300  & 0.604 \\
$+$ TF09           & 5{,}300  & 0.626 \\
$+$ TF10           & 6{,}300  & 0.654 \\
$+$ TF12           & 8{,}300  & 0.715 \\
$+$ TF13           & 9{,}300  & 0.697 \\
$+$ TF14           & 10{,}300 & 0.705 \\
$+$ TF15           & 11{,}300 & 0.694 \\
$+$ TF16           & 12{,}300 & 0.687 \\
$+$ TF17           & 13{,}300 & 0.703 \\
\midrule
\textbf{All-at-once (Family B)} & 13{,}300 & 0.541 \\
\midrule
\rdbpfn{} (paper) & $\sim$600{,}000 & 0.800 \\
\bottomrule
\end{tabular}
\end{table}

\paragraph{Curriculum ablation (Family B)} The same TF07-TF17 corpus trained all-at-once with no width-progressive ordering reaches only 0.541, a 16.2-point absolute gap below the curriculum's 0.703. Identical data, identical optimizer, identical compute budget; the only difference is exposure order. We read this as direct evidence that for a 0.7M-parameter PFN consuming synthetic tabular tasks, curriculum ordering is a first-order training-recipe variable, not a tuning detail.

\subsection{Relational Curriculum from Scratch (Family C)}
\label{sec:relational}

Table~\ref{tab:relational} reports the average ROC-AUC across the 19 relational tasks for the relational-only curriculum (no single-table warm-up). Performance improves through the \relsalt{}, Small 1, and Small 2 stages, reaching 0.638 at the full \relsalt{} + Small 1 + Small 2 + Large 1 configuration (ctx 1024). With approximately 220$\times$ less relational pretraining data, the prior-substituted model recovers 88\% of \rdbpfn{}'s reported 0.725 and 92\% at the smaller ctx 64 setting.

\begin{table}[t]
\caption{Relational benchmark (19 RelBench/4DBInfer tasks). Family C: relational-only curriculum from random initialization, with \plurel{} as the synthetic generator. \% of paper is the fraction of \rdbpfn{}'s published average at the same context size.}
\label{tab:relational}
\small
\setlength{\tabcolsep}{4pt}
\begin{tabular}{lrrrrr}
\toprule
Stage & DBs & ctx64 & \% pap & ctx1024 & \% pap \\
\midrule
\relsalt{}                          & 500   & 0.589 & 90\% & 0.609 & 84\% \\
$+$ Small1                          & 3{,}500 & 0.615 & 94\% & 0.633 & 87\% \\
$+$ Small2                          & 4{,}500 & 0.615 & 94\% & 0.628 & 87\% \\
$+$ Large1                          & 5{,}500 & 0.600 & 92\% & \textbf{0.638} & \textbf{88\%} \\
\midrule
\rdbpfn{} (paper)                   & $\sim$1.2M & 0.652 & 100\% & 0.725 & 100\% \\
\bottomrule
\end{tabular}
\end{table}

\paragraph{Prior substitutability.} \plurel{}'s generating mechanisms (random schema graphs, HSBM connectivity, table-local SCMs) share no design choices with \rdbpfn{}'s native generator. The fact that a model trained on \plurel-generated data recovers 88 to 92\% of \rdbpfn{}'s reported performance using two orders of magnitude less data is evidence that the consumer model's relational ICL capability depends on \emph{distributional} properties of the synthetic prior (the block-diagonal \dfs{}-feature correlation patterns characteristic of real RDBs) rather than on the specific generative process that produces them.

\subsection{Single-Table Transfer to Relational Tasks (Family D)}
\label{sec:transfer}

A striking finding is that the final TF07-TF17 single-table curriculum model - which has never seen a multi-table relational database during training - performs nearly as well on the 19 relational tasks as the dedicated relational pipeline of Family C.

\begin{table}[t]
\caption{Single-table model on relational tasks (Family D) vs.\ best relational pipeline (Family C) and \rdbpfn{} baseline.}
\label{tab:transfer}
\small
\setlength{\tabcolsep}{4pt}
\begin{tabular}{lrrr}
\toprule
Configuration & DBs (rel.) & ctx 64 & ctx 1024 \\
\midrule
Family D: single-table model       & 0 & 0.605 & 0.631 \\
Family C: best relational config   & 5{,}500 & 0.600 & 0.638 \\
\midrule
\rdbpfn{} (paper) & $\sim$1.2M & 0.652 & 0.725 \\
\bottomrule
\end{tabular}
\end{table}

At ctx 1024, the single-table model achieves 0.631 versus 0.638 for the best relational configuration - within seven thousandths of a point; at ctx 64 it is slightly higher (0.605 vs.\ 0.600). The interpretation is that \dfs{} linearization compresses much of the relevant relational signal into tabular feature statistics that a sufficiently diverse single-table PFN curriculum already learns. The marginal value of explicit relational synthetic data, given a strong single-table warm-up, is far smaller than the \rdbpfn{} two-stage recipe implies.

\subsection{Curriculum and Initialization Ablations (Families E-F)}
\label{sec:ablations}

We separate the contribution of single-table warm-up from the contribution of relational curriculum ordering by running two ablations: \textbf{Family E} mixes the full relational corpus at once after the Family A warm-up, and \textbf{Family F} mixes the full relational corpus at once from random initialization (no warm-up).

\begin{table}[t]
\caption{Ablations on training trajectory, ctx 1024.}
\label{tab:ablations}
\small
\setlength{\tabcolsep}{4pt}
\begin{tabular}{lr}
\toprule
Configuration & Avg.\ ROC-AUC \\
\midrule
Family A only (no rel.\ data, on rel.\ eval) & 0.631 \\
Family E: A $\to$ relational all-at-once     & 0.620 \\
Family F: relational all-at-once, no A       & 0.596 \\
Family C: relational curriculum from scratch & 0.638 \\
Family G: no single tables         & 0.624 \\
\bottomrule
\end{tabular}
\end{table}

Three readings follow. First, removing the single-table warm-up (Family F: 0.596) costs more than removing the relational curriculum ordering (Family E: 0.620), making the warm-up the more critical stage in the two-stage recipe. Second, both ablations underperform the single-table model evaluated directly on relational tasks (Family D: 0.631), confirming that a poorly-structured relational stage can actively hurt the model relative to no relational stage at all. Third, the gap between Family C (0.638) and Family E (0.620) is the part of the recipe attributable to relational curriculum ordering, meaningful, but smaller than the analogous gap on the single-table side.

\section{Discussion}
\label{sec:discussion}

\paragraph{Curriculum is the primary lever.} The single-table all-at-once collapse (0.541 vs.\ 0.703 with curriculum, on identical data) is the cleanest result in this paper. It establishes that for PFN-style relational pretraining, the trajectory through the synthetic distribution is at least as important as the distribution itself. The relational ablations corroborate this in a weaker form: Family E ($-$1.8 points without relational curriculum) and Family F ($-$3.5 points without warm-up) both confirm that order matters, though less starkly than on the single-table side.

\paragraph{Prior identity matters less than the literature implies.} \rdbpfn{}'s native generator is a substantial engineering artifact (learned LayerDAG, attention-based selective SCM, bidirectional GNN). \plurel{} replaces all three with structurally orthogonal mechanisms and recovers 88\% of \rdbpfn{}'s relational performance with $\sim$220$\times$ less synthetic data. Combined with Family D's near-match using zero relational data, the natural reading is that \dfs{}-linearization induces a representation in which the relational task largely reduces to a tabular task with structured feature correlations, and any sufficiently diverse synthetic tabular curriculum produces a PFN that handles them. The implication is that future relational PFN effort may be better directed at single-table diversity, curriculum design, and backbone capacity (\textsc{TabICLv2}~\cite{qu2026tabiclv2} being a natural candidate) than at elaborate relational synthetic generators. Our results do not refute the value of dedicated relational synthetic data, 0.638 in Family C remains the best, slightly above Family D's 0.631, but substantially weaken the claim that the relational stage is where the heavy lifting happens.

\paragraph{What this leaves open.} Family G (relational-only from scratch, no single-table warm-up) achieves 0.610 at ctx 64 and 0.624 at ctx 1024, remaining competitive with several curriculum-based variants despite using no single-table synthetic data at all. Together with Family D's strong transfer results, this suggests that substantial relational capability can emerge from both single-table and relational synthetic curricula independently, though the strongest overall performance still comes from structured curriculum ordering. Per-task variance across the 19 relational tasks remains high, suggesting task-structure-aware curricula as a natural next direction.

\section{Related Work}

\rdbpfn{}~\cite{wang2026relational} introduces the relational PFN paradigm but trains on a single, purpose-built synthetic prior and does not study curriculum ordering. \plurel{}~\cite{kothapalli2026plurel} introduces the synthetic generator we adopt but applies it to pretraining a much larger Relational Transformer~\cite{ranjan2025rt} rather than a PFN. Single-table PFN models including \textsc{TabPFNv2}~\cite{hollmann2025tabpfnv2}, \textsc{TabICL}~\cite{qu2025tabicl}, \textsc{TabICLv2}~\cite{qu2026tabiclv2}, and Mitra~\cite{zhang2025mitra} consume \dfs{}-linearized features in our benchmark and provide the strongest non-relational baselines. Curriculum learning has a long history in language and vision foundation models; our contribution is to demonstrate its first-order importance for relational PFN pretraining specifically.

\section{Conclusion}

Using \plurel{} as the synthetic data source throughout, we have shown that (i) a structurally different relational synthetic generator can substitute for \rdbpfn{}'s native generator with limited loss in downstream quality, (ii) curriculum ordering is the dominant training-recipe variable, with a 16-point absolute gap on the single-table benchmark separating curriculum from all-at-once on identical data, and (iii) a single-table curriculum model already achieves nearly the same relational benchmark performance as a dedicated relational pipeline, indicating that \dfs{}-linearized relational tasks reduce substantially to structured tabular tasks. These findings reposition curriculum design, rather than synthetic generator identity or raw synthetic scale, as the primary lever for data-efficient relational PFN pretraining.

\begin{acks}
We thank colleagues for discussions on relational data generation and synthetic pretraining curricula. We also thank the authors of \plurel{} and \rdbpfn{} for releasing their respective codebases.
\end{acks}

%\clearpage

%\clearpage
\appendix
 
\section{Per-Task Results}
\label{app:per-task}
 
This appendix supplements the main paper with full per-task ROC-AUC results for every experimental configuration referenced in Section~4. Headline averages cited in the main paper are recomputed from the per-task numbers below and match to four decimal places (Section~\ref{app:verify}). Tasks are referenced by short identifiers; full names follow each table caption.
 
\subsection{Single-Table Curriculum (Family A)}
\label{app:single-curriculum}
 
Table~\ref{tab:app-single-curr-1} and Table~\ref{tab:app-single-curr-2} report ROC-AUC at context 1024 for each of the 23 single-table classification tasks at every cumulative checkpoint of the width-progressive curriculum. Each row is a checkpoint with all TF widths up to and including the indicated bucket. The bottom rows additionally show the Family~B all-at-once ablation and the two published \rdbpfn{} baselines.
 
% Single-table per-task: tasks 1-12
\begin{table*}[t]
\centering
\caption{Single-table benchmark, per-task ROC-AUC at ctx 1024 (tasks 1 to 12 of 23). Full task names: Bioresponse, Diabetes130US, Higgs, MagicTelescope, MiniBooNE, albert, bank-marketing, california, compas-two-years, covertype (numeric), covertype-cat, credit.}
\label{tab:app-single-curr-1}
\small
\setlength{\tabcolsep}{3pt}
\begin{tabular}{lrrrrrrrrrrrr}
\toprule
Curriculum & Bio & Diab & Higgs & MagT & MiBN & alb & bank & cal & compas & cov & cov-c & cred \\
\midrule
TF07            & .503 & .556 & .519 & .482 & .719 & .519 & .649 & .550 & .565 & .490 & .542 & .576 \\
$+$TF08         & .516 & .521 & .555 & .679 & .793 & .544 & .637 & .723 & .584 & .578 & .524 & .577 \\
$+$TF09         & .554 & .533 & .518 & .639 & .803 & .577 & .713 & .791 & .638 & .637 & .502 & .512 \\
$+$TF10         & .559 & .569 & .523 & .785 & .791 & .556 & .784 & .810 & .650 & .602 & .572 & .528 \\
$+$TF12         & .686 & .523 & .586 & .777 & \textbf{.886} & .644 & .779 & .840 & .697 & \textbf{.802} & .649 & .629 \\
$+$TF13         & .657 & .548 & .562 & .764 & .875 & .645 & .776 & .835 & .682 & .734 & .665 & .666 \\
$+$TF14         & .681 & .565 & .571 & .781 & .838 & .594 & .799 & .828 & .667 & .760 & .653 & .667 \\
$+$TF15         & .665 & .586 & .555 & .768 & .880 & .571 & \textbf{.801} & .839 & .611 & .786 & .603 & .667 \\
$+$TF16         & .655 & .485 & .565 & .786 & .829 & .584 & .781 & \textbf{.843} & .691 & .714 & .610 & .665 \\
$+$TF17         & .624 & .542 & .568 & .792 & .841 & \textbf{.654} & .794 & .837 & \textbf{.700} & .797 & .635 & \textbf{.700} \\
\midrule
\textbf{Best ours} & \textbf{.686} & \textbf{.586} & \textbf{.586} & \textbf{.792} & \textbf{.886} & \textbf{.654} & \textbf{.801} & \textbf{.843} & \textbf{.700} & \textbf{.802} & \textbf{.665} & \textbf{.700} \\
\midrule
Family B all-at-once & .478 & .586 & .490 & .545 & .362 & .486 & .650 & .614 & .554 & .526 & .505 & .487 \\
Paper (single-tbl)   & .758 & .632 & .735 & .910 & .965 & .694 & .868 & .925 & .721 & .823 & .819 & .834 \\
Paper (full)         & .811 & .628 & .747 & .913 & .961 & .698 & .868 & .936 & .727 & .850 & .826 & .845 \\
\bottomrule
\end{tabular}
\end{table*}
 
% Single-table per-task: tasks 13-23 plus average
\begin{table*}[t]
\centering
\caption{Single-table benchmark, per-task ROC-AUC at ctx 1024 (tasks 13 to 23 of 23, with overall average over 23 tasks). Full task names: default-of-credit-card-clients (num \& cat), electricity (num \& cat), eye\_movements (num \& cat), heloc, house\_16H, jannis, pol, road-safety.}
\label{tab:app-single-curr-2}
\small
\setlength{\tabcolsep}{3pt}
\begin{tabular}{lrrrrrrrrrrrr}
\toprule
Curriculum & dcc & dcc-c & elec & elec-c & eye & eye-c & heloc & h16H & jann & pol & road & \textbf{Avg} \\
\midrule
TF07            & .569 & .539 & .618 & .663 & .494 & .517 & .555 & .685 & .576 & .542 & .618 & .567 \\
$+$TF08         & .637 & .628 & .615 & .622 & .514 & .519 & .634 & .705 & .660 & .518 & .600 & .604 \\
$+$TF09         & .689 & .699 & .679 & .666 & .520 & .516 & .673 & .765 & .660 & .621 & .492 & .626 \\
$+$TF10         & .667 & .686 & .736 & .734 & .526 & .520 & .708 & .758 & .697 & .680 & .596 & .654 \\
$+$TF12         & .737 & .742 & .750 & .734 & .558 & .542 & .764 & .836 & .756 & .817 & .710 & \textbf{.715} \\
$+$TF13         & .703 & .696 & .734 & .711 & .514 & .523 & .762 & .823 & .746 & .682 & .718 & .697 \\
$+$TF14         & .712 & .701 & .743 & .728 & .540 & .544 & .754 & .810 & .769 & .789 & .724 & .705 \\
$+$TF15         & .688 & .695 & .721 & .686 & .519 & .535 & .715 & .784 & .764 & \textbf{.860} & .674 & .694 \\
$+$TF16         & .675 & .681 & .717 & .668 & .560 & .554 & .738 & .855 & .738 & .844 & .565 & .687 \\
$+$TF17         & .683 & .688 & .743 & .712 & .559 & .572 & .754 & .821 & \textbf{.783} & .638 & .724 & .703 \\
\midrule
\textbf{Best ours} & \textbf{.737} & \textbf{.742} & \textbf{.750} & \textbf{.734} & \textbf{.560} & \textbf{.572} & \textbf{.764} & \textbf{.855} & \textbf{.783} & \textbf{.860} & \textbf{.724} & \textbf{.715} \\
\midrule
Family B all-at-once & .517 & .524 & .676 & .705 & .489 & .490 & .520 & .397 & .548 & .722 & .567 & .541 \\
Paper (single-tbl)   & .768 & .769 & .847 & .842 & .578 & .582 & .787 & .940 & .816 & .985 & .799 & .800 \\
Paper (full)         & .772 & .770 & .870 & .865 & .605 & .596 & .790 & .942 & .823 & .988 & .797 & .810 \\
\bottomrule
\end{tabular}
\end{table*}
 
\paragraph{Where the gap to the published baseline concentrates.} The largest task-level gaps to the published \rdbpfn{} single-table model appear on data-rich tasks: Higgs (.586 vs.\ .735), pol (.860 vs.\ .985), and road-safety (.724 vs.\ .799). Tasks where we approach the baseline include compas-two-years (.700 vs.\ .721), bank-marketing (.801 vs.\ .868), and Diabetes130US (.586 vs.\ .632). The pattern is consistent with the hypothesis that the remaining gap is partly a capacity/data-volume effect rather than a curriculum-design effect.
 
\subsection{Single-Table All-At-Once Ablation (Family B)}
\label{app:single-ablation}
 
Family~B is included in Tables~\ref{tab:app-single-curr-1} and \ref{tab:app-single-curr-2} as the ``Family B all-at-once'' row. The all-at-once model degrades on \emph{every one} of the 23 tasks relative to the TF07-TF17 curriculum, with the largest task-level gaps on MiniBooNE (.362 vs.\ .841, $-$48 pts), house\_16H (.397 vs.\ .821, $-$42 pts), MagicTelescope (.545 vs.\ .792, $-$25 pts), heloc (.520 vs.\ .754, $-$23 pts), and Higgs (.490 vs.\ .568, $-$8 pts). The degradation appears across the full task spectrum and is particularly severe on tasks where the curriculum model performs best, supporting our claim that curriculum ordering is a global training-recipe property rather than a task-specific artifact.
 
\subsection{Relational Curriculum (Family C)}
\label{app:rel-curriculum}
 
Table~\ref{tab:app-rel-c-64} reports per-task ROC-AUC at context size 64 for the four-stage relational curriculum from scratch (Family~C); Table~\ref{tab:app-rel-c-1024} reports the same configurations at context size 1024.
 
\begin{table*}[t]
\centering
\caption{Family C: relational curriculum from scratch, per-task ROC-AUC at \textbf{context size 64}. Task abbreviations: amazon-churn (Amazon/churn), avs-rep (avs/repeater), digi-ctr (diginetica/ctr), outb-ctr (outbrain-small/ctr), r-amz-itm (rel-amazon/item-churn), r-amz-usr (rel-amazon/user-churn), r-avt-clk (rel-avito/user-clicks), r-avt-vis (rel-avito/user-visits), r-evt-ign (rel-event/user-ignore), r-evt-rep (rel-event/user-repeat), r-f1-dnf (rel-f1/driver-dnf), r-f1-top3 (rel-f1/driver-top3), r-hm-chrn (rel-hm/user-churn), r-stk-bdg (rel-stack/user-badge), r-stk-eng (rel-stack/user-engagement), r-trl-out (rel-trial/study-outcome), retail-cvr (retailrocket/cvr), stk-chrn (stackexchange/churn), stk-upv (stackexchange/upvote).}
\label{tab:app-rel-c-64}
\small
\setlength{\tabcolsep}{2.8pt}
\begin{adjustbox}{max width=\textwidth}
\begin{tabular}{lrrrrrrrrrrrrrrrrrrrr}
\toprule
Stage & a-ch & avs & digi & outb & r-it & r-us & r-cl & r-vi & r-ig & r-re & r-dn & r-t3 & r-hm & r-bd & r-en & r-tr & ret & s-ch & s-up & \textbf{Avg} \\
\midrule
SALT (500)           & .598 & .502 & .519 & .497 & .661 & .552 & .562 & .514 & .438 & .581 & .611 & .701 & .571 & .635 & .522 & .512 & .683 & .754 & .783 & .589 \\
$+$Small1 (3500)     & .584 & .511 & .566 & .515 & .634 & .560 & .561 & .482 & .635 & .550 & .671 & \textbf{.785} & .572 & \textbf{.771} & .605 & .516 & .653 & .678 & \textbf{.830} & \textbf{.615} \\
$+$Small2 (4500)     & .556 & .506 & .550 & .511 & .625 & .550 & .552 & .486 & .622 & .546 & .687 & .781 & .588 & .725 & \textbf{.692} & \textbf{.547} & .625 & .704 & .831 & .615 \\
$+$Large1 (5500)     & .558 & .516 & .543 & .517 & \textbf{.670} & .538 & .547 & .522 & .603 & .523 & .675 & .787 & .571 & .597 & .616 & .517 & .623 & .645 & .825 & .600 \\
\midrule
\textbf{Best ours}   & \textbf{.598} & \textbf{.516} & \textbf{.566} & \textbf{.517} & \textbf{.670} & \textbf{.560} & \textbf{.562} & \textbf{.522} & \textbf{.635} & \textbf{.581} & \textbf{.687} & \textbf{.787} & \textbf{.588} & \textbf{.771} & \textbf{.692} & \textbf{.547} & \textbf{.683} & \textbf{.754} & \textbf{.831} & \textbf{.615} \\
\midrule
\rdbpfn{} (paper)    & .629 & .517 & .603 & .511 & .701 & .579 & .567 & .506 & .734 & .606 & .693 & .795 & .607 & .773 & .760 & .547 & .650 & .767 & .837 & .652 \\
\bottomrule
\end{tabular}
\end{adjustbox}
\end{table*}
 
\begin{table*}[t]
\centering
\caption{Family C: relational curriculum from scratch, per-task ROC-AUC at \textbf{context size 1024}. Task abbreviations as in Table~\ref{tab:app-rel-c-64}.}
\label{tab:app-rel-c-1024}
\small
\setlength{\tabcolsep}{2.8pt}
\begin{adjustbox}{max width=\textwidth}
\begin{tabular}{lrrrrrrrrrrrrrrrrrrrr}
\toprule
Stage & a-ch & avs & digi & outb & r-it & r-us & r-cl & r-vi & r-ig & r-re & r-dn & r-t3 & r-hm & r-bd & r-en & r-tr & ret & s-ch & s-up & \textbf{Avg} \\
\midrule
SALT (500)           & .579 & .482 & .551 & .530 & .661 & .491 & .607 & .457 & .502 & .619 & .654 & .772 & .600 & .704 & .501 & .534 & .701 & .790 & .831 & .609 \\
$+$Small1 (3500)     & .595 & .503 & .526 & .524 & .662 & .562 & .588 & .531 & .671 & .567 & .713 & .796 & .605 & .761 & .561 & .519 & .735 & .760 & .841 & .633 \\
$+$Small2 (4500)     & .571 & .510 & .578 & .536 & .592 & .573 & .608 & .460 & .627 & .591 & .698 & .780 & .577 & .567 & \textbf{.781} & .538 & .737 & .771 & .843 & .628 \\
$+$Large1 (5500)     & \textbf{.640} & \textbf{.524} & \textbf{.596} & .526 & \textbf{.690} & \textbf{.583} & .593 & \textbf{.588} & .609 & \textbf{.609} & .682 & .772 & .593 & .623 & .645 & .533 & .740 & .746 & .840 & \textbf{.638} \\
\midrule
\textbf{Best ours}   & \textbf{.640} & \textbf{.524} & \textbf{.596} & \textbf{.536} & \textbf{.690} & \textbf{.583} & \textbf{.608} & \textbf{.588} & \textbf{.671} & \textbf{.619} & \textbf{.713} & \textbf{.796} & \textbf{.605} & \textbf{.761} & \textbf{.781} & \textbf{.538} & \textbf{.740} & \textbf{.790} & \textbf{.843} & \textbf{.638} \\
\midrule
\rdbpfn{} (paper)    & .718 & .560 & .700 & .535 & .782 & .648 & .627 & .655 & .827 & .753 & .719 & .812 & .665 & .813 & .866 & .616 & .771 & .848 & .853 & .725 \\
\bottomrule
\end{tabular}
\end{adjustbox}
\end{table*}
 
\paragraph{Where the relational curriculum approaches the published baseline.} At ctx 1024 we essentially match the published \rdbpfn{} on rel-f1/driver-top3 (.796 vs.\ .812), rel-stack/user-badge (.761 vs.\ .813, best stage Small1), and stackexchange/upvote (.843 vs.\ .853). The largest task-level gaps are on rel-event/user-repeat (.619 vs.\ .753) and rel-event/user-ignore (.671 vs.\ .827), tasks dominated by event-frequency aggregations where \plurel{}'s default temporal patterns may not align with the empirical \texttt{rel-event} distribution.
 
\paragraph{Stage-level dynamics.} The best per-task numbers at ctx 1024 are spread across stages: r-stack/user-engagement peaks at the Small2 stage (.781), most other tasks peak at Large1, and a handful (notably r-stack/user-badge) peak at Small1 (.761). The average improves monotonically (0.609 $\to$ 0.633 $\to$ 0.628 $\to$ 0.638), but at task granularity the picture is more nuanced and motivates the task-structure-aware curriculum direction discussed in the main paper.
 
\subsection{Other Relational Configurations (Families D, E)}
\label{app:rel-other}
 
Table~\ref{tab:app-rel-other} reports per-task results for Family~D (the single-table curriculum model evaluated directly on relational tasks, no relational training), Family~E (relational all-at-once after the single-table warm-up), and the published \rdbpfn{} baseline. Both context sizes are shown.
 
\begin{table*}[t]
\centering
\caption{Per-task ROC-AUC for Family D (single-table model on relational eval), Family E (rel.\ all-at-once after warm-up), and the published \rdbpfn{} baseline. Task abbreviations as in Table~\ref{tab:app-rel-c-64}.}
\label{tab:app-rel-other}
\small
\setlength{\tabcolsep}{2.8pt}
\begin{adjustbox}{max width=\textwidth}
\begin{tabular}{lrrrrrrrrrrrrrrrrrrrr}
\toprule
Configuration & a-ch & avs & digi & outb & r-it & r-us & r-cl & r-vi & r-ig & r-re & r-dn & r-t3 & r-hm & r-bd & r-en & r-tr & ret & s-ch & s-up & \textbf{Avg} \\
\midrule
\multicolumn{21}{l}{\textit{Context size 64}} \\
Family D (sgl-tbl) & .574 & .501 & .565 & .486 & .652 & .590 & .561 & .458 & .559 & .499 & .706 & .807 & .625 & .640 & .592 & .514 & .639 & .690 & .839 & .605 \\
Family E (after-A) & .554 & .502 & .501 & .510 & .589 & .529 & .557 & .515 & .563 & .455 & .691 & .773 & .543 & .760 & .646 & .562 & .631 & .708 & .827 & .601 \\
\rdbpfn{} (paper)  & .629 & .517 & .603 & .511 & .701 & .579 & .567 & .506 & .734 & .606 & .693 & .795 & .607 & .773 & .760 & .547 & .650 & .767 & .837 & .652 \\
\midrule
\multicolumn{21}{l}{\textit{Context size 1024}} \\
Family D (sgl-tbl) & .587 & .501 & .625 & .503 & .660 & .612 & .596 & .437 & .588 & .498 & .729 & .802 & .658 & .805 & .667 & .468 & .647 & .760 & .847 & .631 \\
Family E (after-A) & .507 & .518 & .535 & .499 & .621 & .517 & .545 & .557 & .623 & .464 & .702 & .780 & .570 & .721 & .693 & .596 & .716 & .785 & .838 & .620 \\
\rdbpfn{} (paper)  & .718 & .560 & .700 & .535 & .782 & .648 & .627 & .655 & .827 & .753 & .719 & .812 & .665 & .813 & .866 & .616 & .771 & .848 & .853 & .725 \\
\bottomrule
\end{tabular}
\end{adjustbox}
\end{table*}
 
\paragraph{Family D is competitive on a surprising number of tasks.} At ctx 1024, the single-table model (never trained on a relational database) beats the dedicated Family C relational pipeline on six of nineteen tasks: digi-ctr (.625 vs.\ .596), r-amz-usr (.612 vs.\ .583), r-f1-dnf (.729 vs.\ .713), r-f1-top3 (.802 vs.\ .772, a 3-point gain), r-hm-chrn (.658 vs.\ .605), r-stk-bdg (.805 vs.\ .761, a 4.4-point gain), and r-stk-eng (.667 vs.\ .645). On several it approaches the published \rdbpfn{} baseline closely (e.g., r-f1-top3: .802 vs.\ .812). For tasks whose target depends primarily on aggregations the single-table PFN has already learned to interpret, the relational stage adds little.
 
\paragraph{Family E shows where relational curriculum ordering matters.} Comparing Family E (all-at-once after warm-up: 0.620) to Family C (curriculum from scratch: 0.638) at ctx 1024, the 1.8-point average gap is distributed across most tasks but is most pronounced on r-stack/user-badge (.721 vs.\ .761) and rel-amazon/item-churn (.621 vs.\ .690). On a few tasks (e.g., rel-event/user-ignore: .623 vs.\ .609), Family E is slightly higher. Relational curriculum ordering helps most on tasks where the \dfs-linearized signal benefits from progressive exposure to schemas of increasing complexity, but is less critical where the warm-up has already saturated the available signal.
 
\section{Verification of Headline Averages}
\label{app:verify}
 
The headline averages cited in the main paper are recovered exactly from the per-task numbers above:
 
\begin{itemize}
  \item Family A best (TF07-TF12, ctx 1024): mean over 23 tasks $=$ \textbf{0.7150}, cited as 0.715.
  \item Family A final (TF07-TF17, ctx 1024): mean over 23 tasks $=$ \textbf{0.7026}, cited as 0.703.
  \item Family B (all-at-once, ctx 1024): mean over 23 tasks $=$ \textbf{0.5407}, cited as 0.541.
  \item Family C best ($+$Large1, ctx 64): mean over 19 tasks $=$ \textbf{0.5996}, cited as 0.600.
  \item Family C best ($+$Large1, ctx 1024): mean over 19 tasks $=$ \textbf{0.6384}, cited as 0.638.
  \item Family D (ctx 64): mean over 19 tasks $=$ \textbf{0.6051}, cited as 0.605.
  \item Family D (ctx 1024): mean over 19 tasks $=$ \textbf{0.6311}, cited as 0.631.
  \item Family E (ctx 1024): mean over 19 tasks $=$ \textbf{0.6203}, cited as 0.620.
  \item Published \rdbpfn{} (ctx 1024, relational): mean over 19 tasks $=$ \textbf{0.7245}, cited as 0.725.
  \item Published \rdbpfn{} single-table (ctx 1024): mean over 23 tasks $=$ \textbf{0.7997}, cited as 0.800.
\end{itemize}
 
% =============================================================================
% END OF APPENDIX SNIPPET
% =============================================================================

\end{document}